\documentclass[11pt,a4paper]{article}
\usepackage[hyperref]{eacl2021}

\usepackage{times}
\usepackage{latexsym}

\usepackage{todonotes}
\usepackage{url}
\usepackage{float}
\usepackage{stfloats}
\usepackage{graphics}
\usepackage{graphicx}
\usepackage{subfigure}
\usepackage{booktabs}
\usepackage{multirow}
\usepackage{tabularx}
\usepackage{amsmath}
\usepackage{amssymb}
\usepackage{enumitem}

\usepackage{microtype}

\aclfinalcopy 

\newcommand*\samethanks[1][\value{footnote}]{\footnotemark[#1]}

\title{Bridging the gap between supervised classification and unsupervised topic modelling for social-media assisted crisis management}

\author{
    Mikael Brunila\thanks{\;\;Equal contribution.}
    \hfill Rosie Zhao\samethanks
    \hfill Andrei Mircea\samethanks
    \hfill Sam Lumley 
    \hfill Renee Sieber
    \\
    McGill University / Montreal, QC, Canada
    \\
    \texttt{\{rosie.zhao,andrei.romascanu,sam.lumley\}@mail.mcgill.ca} \\
    \texttt{mikael.brunila@gmail.com} \\
    \texttt{renee.sieber@mcgill.ca} \\
}

\date{}

\begin{document}
\maketitle

\begin{abstract}
Social media such as Twitter provide valuable information to crisis managers and affected people during natural disasters. Machine learning can help structure and extract information from the large volume of messages shared during a crisis; however, the constantly evolving nature of crises makes effective domain adaptation essential. Supervised classification is limited by unchangeable class labels that may not be relevant to new events, and unsupervised topic modelling by insufficient prior knowledge. In this paper, we bridge the gap between the two and show that BERT embeddings finetuned on crisis-related tweet classification can effectively be used to adapt to a new crisis, discovering novel topics while preserving relevant classes from supervised training, and leveraging bidirectional self-attention to extract topic keywords. We create a dataset of tweets from a snowstorm to evaluate our method's transferability to new crises, and find that it outperforms traditional topic models in both automatic, and human evaluations grounded in the needs of crisis managers. More broadly, our method can be used for textual domain adaptation where the latent classes are unknown but overlap with known classes from other domains.
\end{abstract}

\section{Introduction} \label{sec:intro}
\subsection{Social Media for Crisis Management}\label{sec:intro_crisis_management}
As climate change increases the frequency of extreme weather events and the vulnerability of affected people, effective crisis management is becoming increasingly important for mitigating the negative effects of these crises \citep{keim_building_2008-1}. In the general crisis management literature, social media has been identified as a useful source of information for crisis managers to gauge reactions from and communicate with the public, increase situational awareness, and enable data-driven decision-making \citep{tobias_using_2011-1,alexander_social_2014-1,jin_examining_2014}. 

\subsection{The Need For NLP}
The large volume and noise-to-signal ratio of social media platforms such as Twitter makes it difficult to extract actionable information, especially at a rate suited for the urgency of a crisis. This has motivated the application of natural language processing (NLP) techniques to help automatically filter information in real-time as a crisis unfolds \citep{muhammad_imran_extracting_2013,emmanouil2015big}.

Other work has investigated the use of NLP models to automatically classify tweets into finer-grained categories that can be more salient to crisis managers and affected people in rapidly evolving situations \citep{ragini_empirical_2016,schulz2014evaluating}. 
Training such classification models typically requires large-scale annotated corpora of crisis-related tweets such as that made available by \citet{imran_twitter_2016}, which covers a variety of countries and natural disasters including flooding, tropical storms, earthquakes, and forest fires.

\subsection{Limitations of Current Methods}
Whereas supervised approaches work well for classifying tweets from the same event as their training data, they often fail to generalize to novel events \citep{nguyen_robust_2017}. A novel event may differ from past events in terms of location, type of event, or event characteristics; all of which can change the relevance of a tweet classification scheme.

Various methods of domain adaptation have been suggested for addressing this issue \citep{li_disaster_2018-1,sopova_domain_2017,alrashdi_automatic_2020}. However, this type of supervised classification assumes that relevant classes remain the same from event to event. Probabilistic topic modelling approaches such as Latent Dirichlet Allocation (\textsc{LDA}) can overcome this limitation and identify novel categorizations \citep{blei_latent_2003-1}. Unfortunately, these unsupervised methods are typically difficult to apply to tweets due to issues of document length and non-standard language \citep{hong_empirical_2010}. Furthermore, the categorizations produced by these models can be difficult to interpret by humans, limiting their usefulness \citep{blekanov_ideal_2020}. 

\subsection{Our Contributions}
To address these issues, we propose a method for the unsupervised clustering of tweets from novel crises, using the representations learned from supervised classification. Specifically, we use the contextual embeddings of a pretrained language model finetuned on crisis tweet classification. 

Our method bridges the gap between supervised approaches and unsupervised topic modelling, improving domain adaptation to new crises by allowing classification of tweets in novel topics while preserving relevant classes from supervised training.  Our model is robust to idiosyncrasies of tweet texts such as short document length and non-standard language, and leverages bi-directional self-attention to provide interpretable topic keywords. 

We assess our approach's transferability to novel crises by creating a dataset of tweets from Winter Storm Jacob, a severe winter storm that hit Newfoundland, Canada in January 2020. This event differs significantly from past crisis tweet classification datasets on which we finetune our model, and allows us to evaluate domain adaptation for novel events. We find that our approach indeed identifies novel topics that are distinct from the labels seen during supervised training.

In line with human-centered machine learning principles \citep{ramos_emerging_2019}, we also create a novel human evaluation task aligned with the needs of crisis managers. We find that, with high inter-rater reliability, our model provides consistently more interpretable and useful topic keywords than traditional approaches, while improving cluster coherence as measured by intruder detection. Automated coherence measures further support these findings. Our code and dataset are available at \url{https://github.com/smacawi/bert-topics}.
 
\section{Related Work} \label{sec:relwork}
\subsection{Topic Modelling} \label{sec:relwork_tm}
Topic modelling in a variety of domains is widely studied. Although there are many existing approaches in the literature, most innovations are compared to the seminal Latent Dirichlet Allocation (\textsc{LDA}) model \citep{blei_latent_2003-1}. LDA is a generative probabilistic model that considers the joint distribution of observed variables (words) and hidden variables (topics). While \textsc{LDA} has well-known issues with short text, approaches such as Biterm Topic Modelling (\textsc{BTM}) have been  developed to address these \citep{yan_biterm_2013}. \textsc{BTM} specifically addresses the sparsity of word co-occurrences: whereas \textsc{LDA} models word-document occurrences, \textsc{BTM} models word co-occurrences (`biterms')  across the entire corpus.

\subsection{Clustering} \label{sec:relwork_clstr}
Topic modelling can be distinguished from clustering, where documents are usually represented in a multi-dimensional vector space and then grouped using  vector similarity measures. These representations are typically formed through matrix factorization techniques \citep{levy_neural_2014} that compress words \citep{mikolov_distributed_2013,pennington_glove_2014} or sentences \citep{mikolov_efficient_2013,lau_empirical_2016} into ``embeddings". Recently, language representation models building on Transformer type neural networks \citep{vaswani_attention_2017} have upended much of NLP and provided new, ``contextualized" embedding approaches \citep{devlin_bert_2019, peters_deep_2018}. Among these is the Bidirectional Encoder Representations from Transformers (BERT) model, which is available pretrained on large amounts of text and can be finetuned on many different types of NLP tasks \citep{devlin_bert_2019}. Embeddings from BERT and other Transformer type language models can potentially serve as a basis for both topic modelling \citep{bianchi_pre-training_2020} and clustering \citep{reimers_classification_2019}, but many questions about their usefulness for these tasks remain open.

The use of embeddings for clustering in the field of crisis management has been explored by \citet{demszky_analyzing_2019} using trained GloVe embeddings. \citet{zahera_fine-tuned_nodate} used contextualized word embeddings from BERT to train a classifier for crisis-related tweets on a fixed set of labels. Using BERT to classify tweets in the field of disaster management was also studied by \citet{ma_tweets_nodate} by aggregating labelled tweets from the CrisisNLP and CrisisLexT26 datasets. While the aforementioned work requires data with a gold standard set of labels, our proposed clustering approach using finetuned BERT embeddings is applied in an unsupervised environment on unseen data --- it invites domain expertise to determine an appropriate set of labels specific to the crisis at hand. 

\subsection{Topic Keyword Extraction} \label{sec:relwork_kw}
A significant issue when clustering text documents is how the keywords of each cluster or topic are determined. Unlike standard topic models such as \textsc{LDA}, clustering approaches do not jointly model the distributions of keywords over topics and of topics over documents. In other words, clusters do not contain any obvious information about which keywords should represent the clusters as topics. While the previous generation of embedding models has been leveraged for interpretable linguistic analysis in a wide variety of settings \citep{garg_word_2018,hamilton_cultural_2016,kozlowski_geometry_2019}, the interpretability of language models in general and BERT in particular remains a contested issue \citep{rogers_primer_2020}.

One promising line of research has been on the attention mechanism of Transformer models \citep{bahdanau_attention_2014,jain_attention_2019}. \citet{clark_what_2019} found that the attention heads of BERT contained a significant amount of syntactic and grammatical information, and \citet{lin_open_2019} concluded that this information is hierarchical, similar to syntactic tree structures. \citet{kovaleva_revealing_2019} noted that different attention heads often carry overlapping and redundant information. However, if attention is to be useful for selecting topic keywords, the crucial question is whether it captures semantic information. \citet{jain_attention_2019} found that attention heads generally correlated poorly with traditional measures for feature importance in neural networks, such as gradients, while \citet{serrano_is_2019} showed that attention can ``noisily" predict the importance of features for overall model performance and \citet{wiegreffe_attention_2019} argued that attention can serve plausible, although not faithful explanations of models. To the best of our knowledge, there is no previous work on leveraging attention to improve topic modelling interpretability.

\subsection{Model Coherence} \label{sec:relwork_coherence}
Whether based on clustering or probabilistic approaches, topic models are typically evaluated by their coherence. While human evaluation is preferable, several automated methods have been proposed to emulate that of human performance \citep{lau_machine_2014}.
In both cases, coherence can be thought of formally as a measure of the extent to which keywords in a topic relate to each other as a set of semantically coherent \emph{facts} \citep{roder_exploring_2015,aletras_evaluating_2013,mimno_optimizing_2011}. If a word states a fact, then the coherence of a set of topic keywords can be measured by computing how strongly a word \(W'\) is confirmed by a conditioning set of words \(W^*\). This can be done either directly where \(W'\) is a word in a set of topic words and \(W^*\) are the other words in the same set \citep[e.g.][]{mimno_optimizing_2011}, or indirectly by computing context vectors for both \(W'\) and \(W^*\) and then comparing these \citep[e.g.][]{aletras_evaluating_2013}. In a comprehensive comparison of different coherence measures, \citet{roder_exploring_2015} found that, when comparing the coherence scores assigned by humans to a set of topics against a large number of automated metrics, indirect confirmation measures tend to result in a higher correlation between human and automated coherence scores.

\subsection{Human Evaluation of Topic Models} \label{sec:relwork_human}
While automated coherence measures can be used to rapidly evaluate topic models such as \textsc{LDA}, studies have shown that these metrics can be uncorrelated --- or negatively correlated --- to human interpretability judgements. \citet{chang_reading_2009} demonstrated that results given by perplexity measures differed from that of their proposed `intrusion' tasks, where humans identify spurious words inserted in a topic, and mismatched topics assigned to a document. Tasks have been formulated in previous works to meaningfully enable human judgement when analyzing the topics \citep{chuang_topic_2013, lee_human_2017}.

The latent topics given by these methods should provide a semantically meaningful decomposition of a given corpus. Formalizing the quality of the resulting latent topics via qualitative tasks or quantitative metrics is even less straightforward in an applied setting, where it is particularly important that the semantic meaning underlying the topic model is relevant to its users. Due to our focus on the comparison between the labels in the CrisisNLP dataset and novel topics discovered by our model, we restrict this part of the analysis to nine topics.

Our work is motivated by structuring crude textual data for practical use by crisis managers, guiding \textit{corpus exploration} and efficient information retrieval. It remains difficult to verify whether the latent space discovered by topic models is both interpretable and useful without a gold standard set of labels.

\section{Methodology} \label{sec:method}
In this section we describe the dataset of snowstorm-related tweets we created to evaluate our model's ability to discover novel topics and transfer to unseen crisis events. We then outline the process by which our model learns to extract crisis-relevant embeddings from tweets, and clusters them into novel topics from which it then extracts interpretable keywords. 

\subsection{Snowstorm Dataset} \label{sec:method_data}
On January 18 2020, Winter Storm Jacob hit Newfoundland, Canada.  As a result of the high winds and severe snowfall, 21,000 homes were left without power. A state of emergency was declared in the province as snowdrifts as high as 15 feet (4.6 m) trapped people indoors \citep{erdman_crippling_2020}. 

Following the Newfoundland Snowstorm, we collected 21,797 unique tweets from 8,471 users between January 17 and January 22 using the Twitter standard search API with the following search terms: \texttt{\#nlwhiteout, \#nlweather, \#Newfoundland, \#nlblizzard2020, \#NLStorm2020, \#snowmaggedon2020, \#stormageddon2020, \#Snowpocalypse2020, \#Snowmageddon, \#nlstorm, \#nltraffic, \#NLwx, \#NLblizzard}. Based on past experience with the Twitter API, we opted to use hashtags to limit irrelevant tweets (e.g. searching for \texttt{blizzard} resulted in half the collected tweets being about the video game company with the same name). We filter retweets to only capture unique tweets and better work within API rate limits. We make the dataset publicly available with our code.

\subsection{Finetuned Tweet Embeddings Model}  \label{sec:method_model}
Our proposed approach, Finetuned Tweet Embeddings \textsc{(FTE)} involves training a model with bidirectional self-attention such as \textsc{BERT} \citep{devlin_bert_2019} to generate embeddings for tweets so that these can be clustered using common off-the-shelf algorithms such as K-Means \citep{lloyd_least_1982, elkan_using_2003}. We then combine activations from the model's attention layers with Term frequency-Inverse document frequency (Tf-Idf) to identify keywords for each cluster and improve model interpretability.

\subsubsection{Model Finetuning} \label{sec:method_model_ft}
To build a model that extracts tweet embeddings containing information relevant to crisis management, we finetune a pretrained \textsc{BERT} language representation model on classifying tweets from various crisis events. Similar to \citet{ourpaper2020}, we finetune on CrisisNLP, a dataset of Crowdflower-labeled tweets from various types of crises, aimed at crisis managers \citep{imran_twitter_2016}. These show a significant class imbalance with large amounts of tweets shunted into uninformative categories such as \texttt{Other useful information}, further motivating the need for unsupervised topic discovery. CrisisNLP label descriptions and counts are included in Appendix \ref{apdx:crisisnlp} for context.  To address the issue of class imbalance, we create a random stratified train-validation split of $0.8$ across the datasets, preserving the same proportions of labels. 

To finetune \textsc{BERT}, we add a dropout layer and a linear classification layer on top of the \texttt{bert-base-uncased} model, using the 768-dimensional \texttt{[CLS]} last hidden state as input to our classifier. We train the model using the Adam optimizer with the default fixed weight decay, and a batch size of four over a single epoch. Our model obtains an accuracy of 0.78 on the withheld validation dataset. One advantage of \textsc{BERT} is its subword tokenization which can dynamically build representations of out-of-vocabulary words from subwords, allowing a robust handling of the non-standard language found in tweets.

\subsubsection{Tweet Embedding} \label{sec:method_model_embed}
Once the model is trained, we use a mean-pooling layer across the last hidden states to generate a tweet embedding, similar to \citet{reimers_sentence-bert_2019} who found mean-pooling to work best for semantic similarity tasks. Whereas the hidden state for the \texttt{[CLS]} token contains sufficient information to separate tweets between the different CrisisNLP labels, the hidden states for the other tokens in the tweet allow our model to capture token-level information that can help in identifying novel topics beyond the supervised labels. For example, tweets that use similar words --- even those not occurring in the CrisisNLP dataset --- will have more similar embeddings and thus be more likely to cluster in the same topic. 

\subsubsection{Clustering} \label{sec:method_model_clstr}
Given the embeddings for each tweet, we apply an optimized version of the K-Means clustering algorithm to find our candidate topics \citep{elkan_using_2003}.  We use the K-Means implementation in \texttt{Sklearn} with the default `k-means++' initialization and n\verb!_!int equal to 10.

\subsubsection{Keyword Extraction} \label{sec:method_model_kw}
To extract keywords from topics generated by our model and ensure their interpretability, we experiment with two approaches and their combination.

We first identify relevant keywords for each cluster using Tf-Idf \citep{sparck_jones_statistical_2004}, combining each cluster into one document to address the issue of low term frequencies in short-text tweets. During automatic evaluation, we perform a comprehensive grid search over Tf-Idf and other hyperparameters: 
\begin{enumerate}[topsep=0pt,itemsep=-1ex,partopsep=1ex,parsep=1ex]
    \item maximum document frequency ($mdf$) between $0.6$ and $1.0$ with intervals of $0.1$ (to ignore snowstorm related terms common to many clusters); 
    
    \item sublinear Tf-Idf (shown to be advantageous by \citet{paltoglou_study_2010}); 
    
    \item phrasing (grouping of frequently co-occurring words proposed by \citet{mikolov_distributed_2013})
\end{enumerate}
We find an $mdf$ of 0.6 and sublinear Tf-Idf perform best for \textsc{FTE} with our number of topics, and we use these hyperparameters in our experiments. Phrasing makes no significant difference and we only include unigrams in our keywords.

However, Tf-Idf only uses frequency and does not leverage the crisis-related knowledge learned during finetuning. Based on the observation by \citet{clark_what_2019}, we use \textsc{BERT}'s last layer of attention for the \texttt{[CLS]} token to identify keywords that are important for classifying tweets along crisis management related labels. For each cluster, we score keywords by summing their attention values (averaged across subwords) across tweets where they occur, better capturing the relevance of a keyword to crisis management. 

We also experiment with the combination of Tf-Idf and attention by multiplying the two scores for each token, allowing us to down-weight frequent but irrelevant words and up-weight rarer but relevant words. For all three approaches, we drop stopwords, hashtags, special characters, and URLs based on preliminary experiments that found these contributing substantially to noise in the topic keywords. 

\section{Evaluation} \label{sec:method_eval}
In this section we describe the baselines, as well as the automatic and human evaluations used.

\subsection{Baselines} \label{sec:method_eval_baselinesl}
 We report on two standard topic modelling techniques: \textsc{BTM} and \textsc{LDA}, for which we train models ranging from five to fifteen topics under 10 passes and 100 iterations, following the work of \citet{blei_latent_2003-1}. For \textsc{LDA}, we generate clusters of tweets by giving a weighted topic assignment to each word present in a given document according to the topic distribution over all words present in the corpus. Clusters can be generated similarly with \textsc{BTM}, but according to the topic distribution over biterms (with a window size of 15). 
 
 We also report on \textsc{BERT}, which is simply our method without the finetuning step, i.e. using vanilla pretrained \textsc{BERT} embeddings with K-Means clustering and keyword extraction based on Tf-Idf and attention. For further comparison with the \textsc{FTE} model, we focus our analysis on trained baselines with nine topics, the number of labels in the CrisisNLP dataset \citep{imran_twitter_2016}.

\subsection{Automatic Evaluation} \label{sec:method_eval_automatic}
To evaluate topics we calculate $C_{NPMI}$ and $C_V$, two topic coherence metrics based on direct and indirect confirmation respectively \citep{roder_exploring_2015}. These are described in Appendix \ref{apdx:cv_cnpmi}. For subsequent human evaluation, we select the configuration (\S \ref{sec:method_model_kw}) of each model with the highest $C_V$ for nine topics. This allows us to directly compare with the nine labels from the CrisisNLP dataset and assess our model's ability to learn novel topics. We focus on $C_V$ as \citet{roder_exploring_2015} found it to correlate best with human judgements (in contrast to \(UMASS\), another commonly used coherence metric that was not included in our analysis due to poor correlation with human judgements).

\subsection{Human Evaluation} \label{sec:method_eval_human}
We performed anonymous evaluation through four annotators\footnote{Student researchers familiar with the crisis management literature and the needs of crisis managers as described in \S \ref{sec:intro_crisis_management}}. Since these models are primarily for use by crisis managers, we aim to concentrate our evaluation from the perspective of annotators working in the field. Specifically, we propose two evaluation methods focused on (1) topic keywords and (2) document clustering within topics.

\subsubsection{Keyword Evaluation} \label{sec:method_eval_human_keyword}
To assess the quality of topic keywords, annotators were presented with the top 10 keywords for each topic (Table \ref{table:topic_keywords}) and asked to assign an interpretability score and a usefulness score on a three-point scale. Following the criteria of \citet{rosner_evaluating_2014}, we define interpretability as \textbf{good} (eight to ten words are related to each other), \textbf{neutral} (four to seven words are related), or \textbf{bad} (at most three words are related). Usefulness in turn considers the ease of assigning a short label to describe a topic based on its keywords, similar to \citet{newman_automatic_2010} except we further require the label should be useful for crisis managers. We score usefulness on a three-point scale: \textbf{useful}, \textbf{average}, or \textbf{useless}. 

\subsubsection{Cluster Evaluation} \label{sec:method_eval_human_cluster}
The second task assesses --- from the perspective of a crisis manager --- the interpretability and usefulness of the actual documents clustered within a topic, instead of only analyzing topic keywords as done in previous work. 

Given an anonymized model, for each topic we sample 10 sets of four documents within its cluster along with one document --- the `intruder' --- outside of that topic. For each set of documents, all four annotators were tasked with identifying the intruder from the sample of five documents, as well as assigning an interpretability score and a usefulness score to each sample.

The task of intrusion detection is a variation of \citet{chang_reading_2009}. However, instead of intruder topics or topic words, we found that assessing intruder tweets would give us a better sense of the differences in the clusters produced by our models. Participants were also given the option of labeling the intruder as `unsure' to discourage guessing.

The interpretability score was graded on a three-point scale: \textbf{good} (3-4 tweets seem to be part of a coherent topic beyond “snowstorm”), \textbf{neutral}, and \textbf{bad} (no tweets seem to be part of a coherent topic beyond “snowstorm”). The cluster usefulness score was similar to the keyword usefulness score, but formulated as a less ambiguous binary assignment of \textbf{useful} or \textbf{useless} for crisis managers wanting to filter information during a crisis. 

\subsection{Model agreement} \label{sec:method_eval_agreement}

While our human cluster evaluation provides a good estimate of how topic clusters appear to humans, it does not necessarily establish the difference between two models' document clusters due to the random sampling involved. In other words, two models may have different cluster evaluation results, but similar topic clusters. We define `agreement' as a measure of the overlap between two unsupervised classification models. 

Given a model $A$, its agreement $\operatorname{Agr}_A$ with a model $B$ is
\begin{equation}
    \operatorname{Agr}_A (B) = \frac{\sum\limits_{i=0}^{N} \max\limits_{j} \operatorname{p}(A_{i}, B_{j})}{N}
\end{equation}

\noindent where $A_i$ is the set of documents in the $i^{th}$ cluster of $A$ and $\operatorname{p} (A_i,B_j)$ is the proportion of documents in $A_i$ that are also in $B_j$. We further define model agreement between $A$ and $B$ as the  average of $\operatorname{Agr}_A(B)$ and $\operatorname{Agr}_B(A)$.

\section{Results} \label{sec:results}

\subsection{Automatic Evaluation} \label{sec:results_automatic}
\begin{figure}[]
    \centering
    \includegraphics[width=\columnwidth]{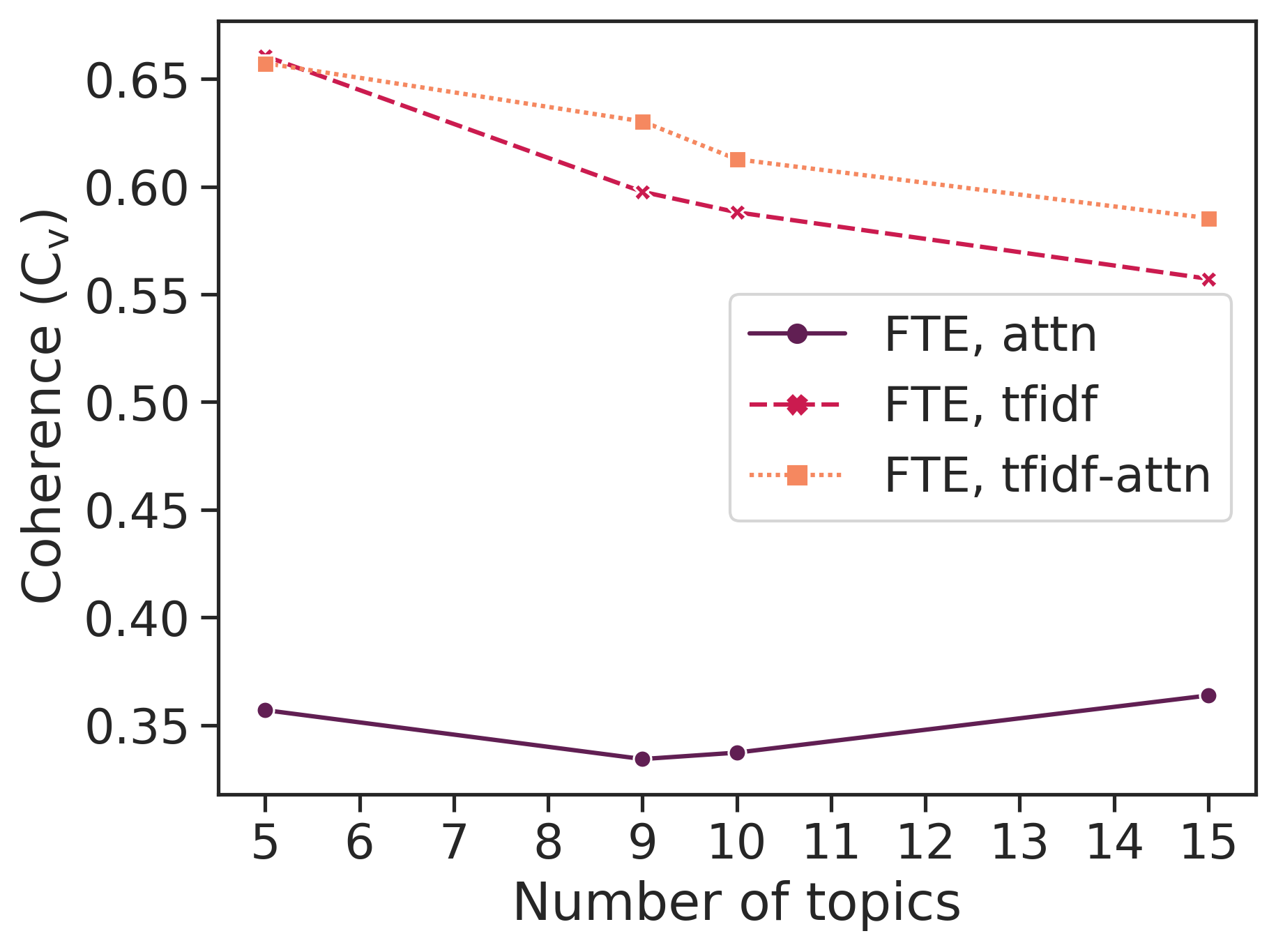}
    \caption{Effect of keyword extraction strategies. 
    }
    \label{fig:coherence_fte}
\end{figure}

Figure \ref{fig:coherence_fte} shows that combining attention and Tf-Idf produces the highest automated \(C_{V}\) coherence scores for our method, across a range of topic numbers. However, the improvement of adding attention is marginal and attention alone performed much worse, suggesting it is suboptimal for identifying keywords. Recent work by \citet{kobayashi-etal-2020-attention} proposes a norm-based analysis which may improve upon this.

\begin{figure}[]
    \centering
    \includegraphics[width=\columnwidth]{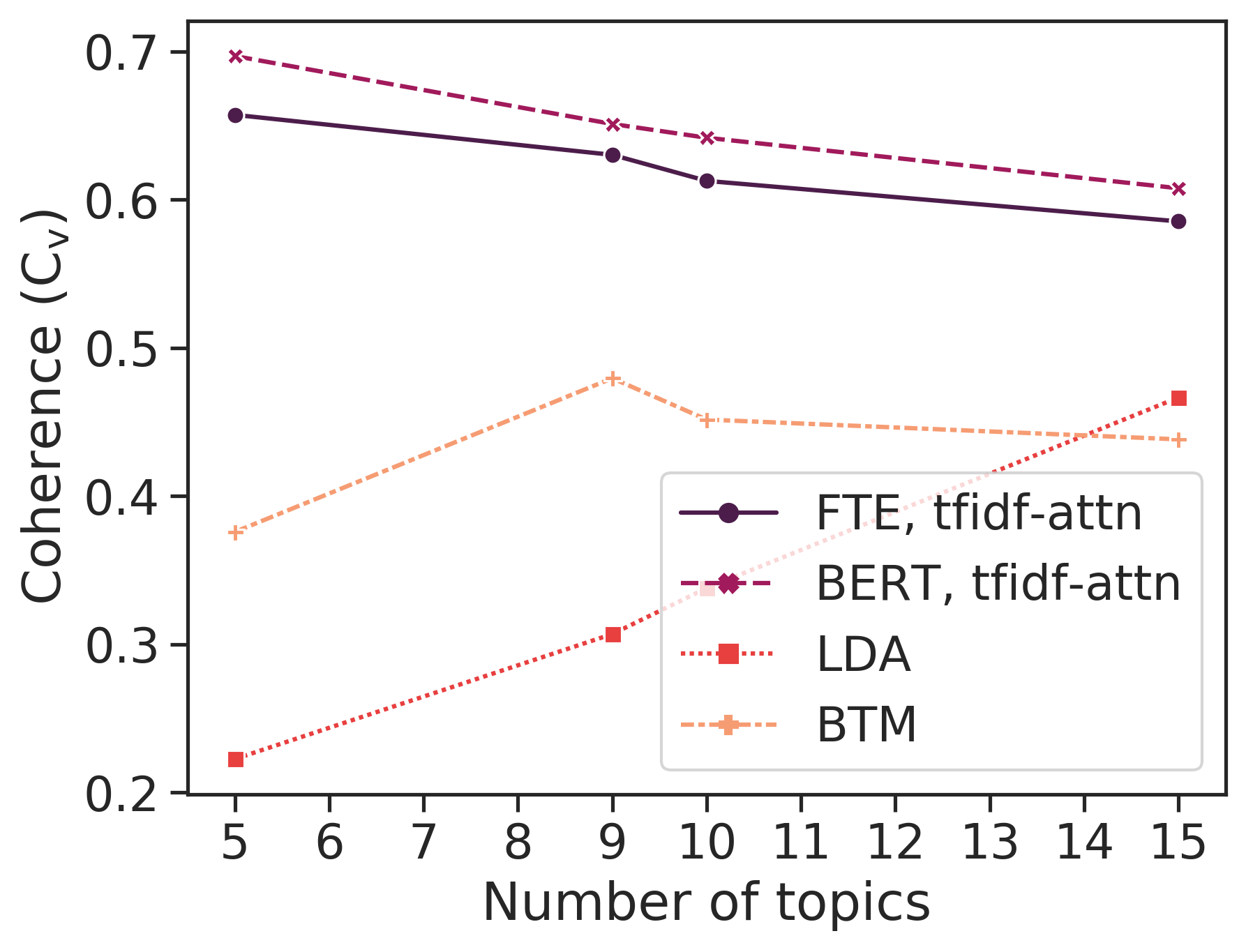}
    \caption{Comparison of \textsc{FTE} with baselines. 
    }
    \label{fig:coherence}
\end{figure}

Figure \ref{fig:coherence} shows that \textsc{FTE} significantly outperforms the \textsc{LDA} and \textsc{BTM} baselines, with similar scores to the \textsc{BERT} baseline. We observed similar trends for \(C_{NPMI}\). However, despite \textsc{BERT}'s high $C_V$ scores, we found that the topics it generated were of very low quality, as described below. 

\subsection{Qualitative Analysis of Keywords} \label{results_qualitative}
Topic keywords are shown in Table \ref{table:topic_keywords}. By restricting the number of topics to the number of labels in the CrisisNLP dataset, we were able to ask our annotators to identify overlap with these original labels. Conversely, this allows us to show that our approach indeed bridges the gap between supervised classification and unsupervised topic modelling by identifying novel topics in addition to salient topics from supervised training. 

\subsubsection{FTE Keywords}
Annotators identified overlap between generated topics and relevant classes from the CrisisNLP dataset: Topic 4 capturing donation needs and volunteering services, Topic 5 expressing sympathy and emotional support, Topic 7 covering missing or trapped people, and Topic 2 seemingly covering unrelated information. 
Distinct novel topics were also identified in the meteorological information in Topic 1, and information about power outages and closures in Topic 3. 
Topics 8 and 9 were less clear to annotators, but the former seemed to carry information about how extreme the storm was thought to be and the latter about citizens bundling up indoors with different foods and activities.

\begin{table*}[t]
    \centering
    \resizebox{\textwidth}{!}{%
    \begin{tabular}{l c c c c c c c c c}
    \toprule
    & \multicolumn{9}{c}{Topic} \\
    Model & 1 & 2 & 3 & 4 & 5 & 6 & 7 & 8 & 9 \\
    \midrule
     \textsc{FTE} & reporting & ivyparkxadidas & outage & assistance & prayer & blowingsnow & trapped & monster & bread \\
     & monster & mood & campus & assist & praying & alert & stranded & meteorologist & song \\
     & snowiest & song & widening & troop & pray & advisory & hydrant & drifting & coffee \\
     & recorded & blackswan & advisory & volunteer & wish & caution & ambulance & perspective & milk \\
     & peak & le & reported & providing & wishing & advised & dead & stormofthecentury & feelin \\
     & temperature & snowdoor & impassable & relief & humanity & stormsurge & garbage & mood & pin \\
     & cloudy & perspective & remaining & aid & brave & wreckhouse & rescue & snowdrift & enjoying \\
     & reported & ode & thousand & request & surviving & surge & permitted & climate & laugh \\
     & equivalent & music & reporting & offering & loved & drifting & body & windy & favorite \\
     & meteorologist & adidasxivypark & suspended & rescue & kindness & avoid & helped & snowdoor & girl \\
    \midrule
     \textsc{BTM} & emergency & eminem & safe & cbcnl & people & like & nltraffic & closed & cm \\
     & state & photo & stay & today & today & storm & road & st & st \\
     & st & click & blizzard & thank & storm & snow & power & tomorrow & winds \\
     & city & learn & newfoundland & people & need & time & st & john & pm \\
     & cityofstjohns & saveng & storm & day & like & day & street & remain & today \\
     & says & michelleobama & canada & home & day & newfoundland & drive & january & km \\
     & declared & sexeducation & weather & work & grocery & going & pearl & update & airport \\
     & john & ken & nlstorm & help & food & blizzard & roads & emergency & yyt \\
     & mayor & starr & warm & storm & open & house & line & state & snowfall \\
     & roads & pin & snowstorm & nltraffic & know & today & mount & today & blizzard \\
    \midrule
     \textsc{BERT} 
        & shareyourweather & pin & thankful & lay & metro & mood & glad & thankful & monster \\
        & ivyparkxadidas & feelin & bank & justintrudeau & campus & hutton & looked & glad & stormsurge \\
        & saturdaythoughts & taxi & yo & save & provincial & cute & apartment & sharing & le \\
        & saturdaymotivation & mayor & mayor & suck & mayor & compound & cloudy & favourite & wreckhouse \\
        & snowdoor & metro & neighbor & radiogregsmith & remain & lovely & law & shareyourweather & newfoundlandlabrador \\
        & bcstorm & en & walked & kettle & operation & spread & snowblowing & grateful & ode \\
        & titanscollections & blowingsnow & wa & anthonygermain & pharmacy & stream & honestly & monster & snowiest \\
        & badboysforlife & ivyparkxadidas & bus & kilbride & region & design & as & feelin & ottawa \\
        & bingo & bus & glad & kaylahounsell & taxi & ivyparkxadidas & eat & neighbor & historic \\
        & blackswan & dannybreennl & pharmacy & campus & advisory & crisis & weird & mom & explorenl \\

    \bottomrule
    \end{tabular}
    }
\caption{Topic keywords for our \textsc{FTE} model and  the \textsc{BTM} and \textsc{BERT} baselines used in human evaluation.}
\label{table:topic_keywords}
\end{table*}

\subsubsection{BTM Keywords}
The topics in \textsc{BTM} were less semantically meaningful to annotators, although they found interesting topics there as well, with Topic 5 showing information about the need to stockpile provisions, Topic 7 relating to traffic conditions, and Topic 8 potentially providing information about a state of emergency and closed businesses.

\subsubsection{BERT Keywords}
The topics in \textsc{BERT} were largely incoherent for annotators, with the exceptions being Topic 8 (positive sentiment) and Topic 5 (services and advisories). This is in stark contrast to the large automated coherence scores obtained by this method, indicating the importance of pairing automatic evaluation with human evaluation. 

\subsection{Human evaluation} \label{sec:results_human}
    \begin{figure*}[t]
      \centering
      \subfigure[Keyword Interpretability]{\includegraphics[width=.48\textwidth]{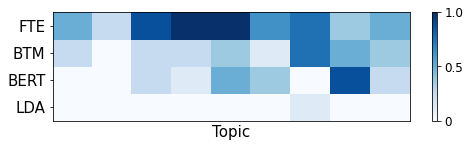} \label{sub-fig:kw_interpret}} \hfill
      \subfigure[Keyword Usefulness]{\includegraphics[width=.48\textwidth]{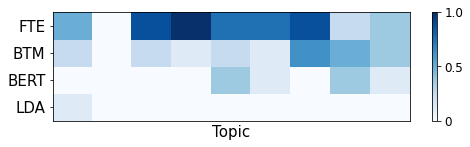} \label{sub-fig:kw_useful}} \hfill
      
      \subfigure[Cluster Interpretability]{\includegraphics[width=.48\textwidth]{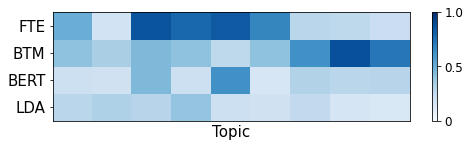} \label{sub-fig:clstr_coherence}} \hfill
      \subfigure[Cluster Usefulness]{\includegraphics[width=.48\textwidth]{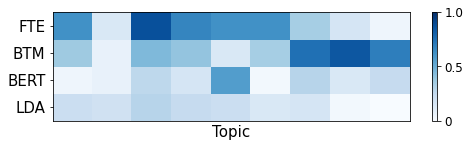} \label{sub-fig:clstr_usefulness}} \hfill

      \subfigure[Correct Intruders]{\includegraphics[width=.48\textwidth]{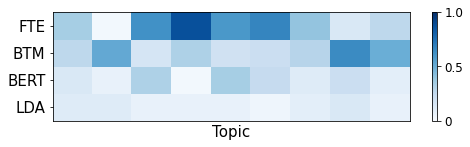} \label{sub-fig:correct_intruders}} 
      \hfill
      \subfigure[Unsure Intruders]{\includegraphics[width=.48\textwidth]{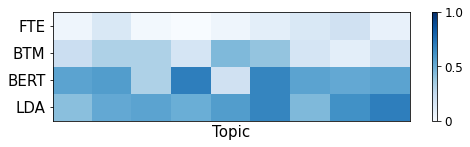} \label{sub-fig:unknown_intruders}} 
      \hfill
    \caption{Topic-level results for keyword and cluster evaluations, aligned with topics from Table \ref{table:topic_keywords}. All scores are rescaled to values between 0 and 1, then averaged across annotators and samples.}
    \label{fig:cluster_scores}
    \end{figure*}

\begin{table}[t]
    \centering
    \resizebox{\columnwidth}{!}{%
    \begin{tabular}{l c c c c c c }
    \toprule
           & \multicolumn{2}{c}{Average Score} & 
           \multicolumn{2}{c}{Topic Count} & 
           \multicolumn{2}{c}{Fleiss'  $\kappa$} \\
     Score &  \textsc{BTM} &  \textsc{FTE} &
      \textsc{BTM} &  \textsc{FTE} &
      \textsc{BTM} &  \textsc{FTE}\\ 
    \midrule
    Interpretability & 31.94  & \textbf{65.28} & 
                         1 &  \textbf{5} &
                         15.01 & \textbf{17.97}\\ 
    Usefulness &  27.78  & \textbf{59.72} & 
                         1 &  \textbf{5} &
                         12.36 & \textbf{21.55}\\ 
    \bottomrule
    \end{tabular}
    }
\caption{Keyword Evaluation scores averaged across topics, number of topics with average scores greater than $0.5$, and inter-rater agreements (Fleiss' $\kappa$).}
\label{table:human_eval_keywords}
\end{table}

\begin{table}[t]
    \centering
    \resizebox{\columnwidth}{!}{%
    \begin{tabular}{l c c c c c c}
    \toprule
           & \multicolumn{2}{c}{Average Score} & \multicolumn{2}{c}{Topic Count} & \multicolumn{2}{c}{Fleiss'  $\kappa$} \\
     Score & \textsc{BTM} &  \textsc{FTE} & \textsc{BTM} &  \textsc{FTE} & \textsc{BTM} &  \textsc{FTE}\\ 
    \midrule
    Interpretability &  50.28 & \textbf{51.53} & 3  & \textbf{4} &  11.05 & \textbf{23.45}\\ 
    Usefulness & 45.46 & \textbf{46.11} & 3 & \textbf{5} & \textbf{21.82} & 21.60\\ 
    Correct Intruders & 35.28 & \textbf{44.17} & 2 & \textbf{4} & 25.78 & \textbf{31.50}\\ 
    Unknown Intruders & 26.39 & \textbf{8.89} & 0 & 0 & - & - \\ 
    \bottomrule
    \end{tabular}
    }
\caption{Cluster Evaluation scores averaged across topics, number of topics with average scores greater than $0.5$, and inter-rater agreements (Fleiss' $\kappa$).}
\label{table:human_eval_cluster}
\end{table}

In Figure \ref{fig:cluster_scores}, we compare topic-level results for keyword evaluations (averaged across annotators), and for cluster evaluations (averaged across samples and annotators). We summarize these results for \textsc{BTM} and \textsc{FTE} in Table \ref{table:human_eval_keywords} and Table \ref{table:human_eval_cluster}, leaving out \textsc{LDA} and \textsc{BERT} due to  significantly lower scores. 

We find that our method outperforms the various baselines on both the keyword and cluster evaluations. In particular, the improvement over \textsc{BERT} further confirms the importance of the finetuning step in our method. In contrast, the improvements in average scores over \textsc{BTM} reported in Table \ref{table:human_eval_cluster} are marginal. Nevertheless, we find that the number of interpretable and useful topic clusters was greater for our approach. Indeed, while the \textsc{BTM} baseline had more semi-interpretable (i.e. only a subset of the sampled tweets seemed related) but non-useful topics, our method had a much clearer distinction between interpretable/useful and non-interpretable/non-useful topics, suggesting that tweets marked as hard to interpret and not useful are consistently irrelevant. This may be preferable for downstream applications, as it allows users to better filter our irrelevant content. 

The annotators also identified intruder tweets in topic samples from \textsc{FTE} more reliably and with less uncertainty, as measured by the number of correct intruders predicted and the number of times an intruder could not be predicted. Interestingly, \textsc{BTM} topics rated for high interpretability had lower rates of correct intruder detection, suggesting that these topics may seem misleadingly coherent to annotators. Inter-rater agreements as measured by Fleiss' $\kappa$ further confirm that annotators more often disagreed on intruder prediction and interpretability scoring for \textsc{BTM} topics. This is undesirable for downstream applications, where poor interpretability of topics can lead to a misinterpretation of data with real negative consequences. 

\subsection{Model agreement} \label{sec:results_agreement}

Agreement for the models was 32.7\% between \textsc{FTE} and \textsc{BERT}, 26.5\% between \textsc{FTE} and \textsc{BTM}, 19.9\% between \textsc{FTE} and \textsc{LDA} and 20.7\% between  \textsc{BTM} and \textsc{LDA}. This confirms that the different models also generate different clusters.

\section{Conclusion} \label{sec:conclusion}
This paper introduces a novel approach for extracting useful information from social media and assisting crisis managers. We propose a simple method that bridges the gap between supervised classification and unsupervised topic modelling  to address the issue of domain adaptation to novel crises.

Our model (\textsc{FTE}, Finetuned Tweet Embeddings) incorporates crisis-related knowledge from supervised finetuning while also being able to generate salient topics for novel crises. To evaluate domain adaptation, we create a dataset of tweets from a crisis that significantly differs from existing crisis Twitter datasets: the 2020 Winter Storm Jacob.

Our paper also introduces human evaluation methods better aligned with downstream use cases of topic modelling in crisis management, emphasizing human-centered machine learning. In both these human evaluations and traditional automatic evaluations, our method outperforms existing topic modelling methods, consistently producing more coherent, interpretable and useful topics for crisis managers. Interestingly, our annotators reported that several coherent topics seemed to be composed of related subtopics. In future work, the number of topics as a hyper-parameter could be explored to see if our approach captures these salient subtopics.

Our method, while simple, is not specific to crisis management and can be more generally used for textual domain adaptation problems where the latent classes are unknown but likely to overlap with known classes from other domains.

\section*{Acknowledgements} \label{sec:acknowledgements}

We are grateful for the funding from Environment and Climate Change Canada (ECCC GCXE19M010). Mikael Brunila also thanks the Kone Foundation for their support.

\bibliographystyle{acl_natbib}
\bibliography{smacawi, smacawi_2}

\appendix
\clearpage

\section{CrisisNLP Dataset} \label{apdx:crisisnlp}
Here we include class distributions (Table \ref{table:crisisnlp_counts}) and descriptions (Table \ref{table:crisisnlp_labels}) from the  CrisisNLP Dataset.

\begin{table}[h]
    \centering
    \resizebox{\columnwidth}{!}{%
    \begin{tabular}{l c c c c c c c c c}
    \toprule
    & \multicolumn{9}{c}{Label Id} \\
    Crisis Dataset & 1 & 2 & 3 & 4 & 5 & 6 & 7 & 8 & 9 \\
    \midrule
    2013\_pak\_eq & 351 & 5 & 16 & 29 & 325 & 75 & 112 & 764 & 336 \\
    2014\_cali\_eq & 217 & 6 & 4 & 351 & 83 & 84 & 83 & 1028 & 157 \\
    2014\_chile\_eq & 119 & 6 & 63 & 26 & 10 & 250 & 541 & 634 & 364 \\
    2014\_odile & 50 & 39 & 153 & 848 & 248 & 77 & 166 & 380 & 52 \\
    2014\_india\_floods & 959 & 14 & 27 & 67 & 48 & 44 & 30 & 312 & 502 \\
    2014\_pak\_floods & 259 & 117 & 106 & 94 & 529 & 56 & 127 & 698 & 27 \\
    2014\_hagupit & 66 & 8 & 130 & 92 & 113 & 349 & 290 & 732 & 233 \\
    2015\_pam & 143 & 18 & 49 & 212 & 364 & 93 & 95 & 542 & 497  \\
    2015\_nepal\_eq & 346 & 189 & 85 & 132 & 890 & 35 & 525 & 639 & 177 \\
    \midrule
    Total & 2510 & 402 & 633 & 1851 & 2610 & 1063 & 1969 & 5729 & 2345\\
    \bottomrule
    \end{tabular}
    }
\caption{Label counts for the different datasets labeled by Crowdflower workers in \citep{imran_twitter_2016}}
\label{table:crisisnlp_counts}
\end{table}

\begin{table*}[b!]
    \centering
    \resizebox{\textwidth}{!}{%
    \begin{tabularx}{\linewidth}{p{0.35\linewidth} c X}
    \toprule
    Label & Id & Description \\
    \midrule
    Injured or dead people  & 
    1 &
    Reports of casualties and/or injured people due to the crisis
    \\
    
    Missing, trapped, or found people & 
    2 &
    Reports and/or questions about missing or found people
    \\
    
    Displaced people and evacuations  &
    3 &
    People who have relocated due to the crisis, even for a short time (includes evacuations
    \\
    
    Infrastructure and utilities damage  &
    4 &
    Reports of damaged buildings, roads, bridges, or utilities/services interrupted or restored
    \\
    
    Donation needs or offers or volunteering services  &
    5 &
    Reports of urgent needs or donations of shelter and/or supplies such as food, water, clothing, money, medical supplies or blood; and volunteering services
    \\
    
    Caution and advice  &
    6 &
    Reports of warnings issued or lifted, guidance and tips
    \\
    
    Sympathy and emotional support  &
    7 &
    Prayers, thoughts, and emotional support
    \\
    
    Other useful information  &
    8 &
    Other useful information that helps one understand the situation
    \\
    
    Not related or irrelevant  &
    9 &
    Unrelated to the situation or irrelevant
    \\
    \bottomrule
    \end{tabularx}
    }
\caption{Label descriptions and id's in \citep{imran_twitter_2016}}
\label{table:crisisnlp_labels}
\end{table*}

\section{Automated Coherence Metrics}  \label{apdx:cv_cnpmi}
The direct confirmation \(C_{NPMI}\), uses a token-by-token ten word sliding window, where each step determines a new virtual document. Co-occurrence in these documents is used to compute the normalized pointwise mutual information (NPMI) between a given topic keyword \(W'\) and each member in the conditioning set of other topic keywords \(W^{*}\), such that:

\begin{align*}
NPMI = (\dfrac{PMI(W', W^{*})}{-log(P(W', W^{*}) + \epsilon)})^{\gamma}
\end{align*}

\begin{align*}
PMI = log\dfrac{P(W', W^{*}) + \epsilon}{P(W')* P(W^{*})}
\end{align*}

\noindent 
The coherence of a topic is then calculated by taking the arithmetic mean of these confirmation values, with \(\epsilon \) as a small value for preventing the log of zero.

The indirect confirmation \(C_{V}\) is instead based on comparing the contexts in which \(W'\) and \(W^{*}\) appear. \(W'\) and \(W^{*}\) are represented as vectors of the size of the total word set \(W\). Each value in these vectors consist of a direct confirmation between the word that the vector represents and the words in \(W\). However, now the context is just the tweet that each word appears in. The indirect confirmation between each word in the topic is the cosine similarity of each pair of context vectors such that
\begin{align*}
    \cos(\vec{u},\vec{w}) = \dfrac{\sum_{i=1}^{|W|}u_{i}\cdot{}u_{i}} {||\vec{u}||_{2}||\vec{w}||_{2}}
\end{align*}

\noindent 
where \(\vec{u} = \vec{v}(W')\) and  \(\vec{w} = \vec{v}(W^*)\). Once again, the arithmetic mean of these similarity values gives the coherence of the topic. 

\(C_{V}\) was found by \citet{roder_exploring_2015} to be the most interpretable topic coherence metric when compared to human judgement and has later been used extensively on assessing the coherence of short texts like tweets as well \citep{zeng_topic_2018,habibabadi_topic_2019,wallner_tweeting_2019}. We also use \(C_{NPMI}\), which has been one of the most successful topic coherence measures based on direct confirmation \citep{roder_exploring_2015,lau_machine_2014}. See \citet{roder_exploring_2015} for further details on both \(C_{V}\) and \(C_{NPMI}\).

\clearpage
\section{Human evaluation tweet samples} \label{apdx:tweet_samples}
Here we present one of the set of tweets presented to human annotators for each model and topic. We also show the ground truth intruder tweet which the annotators were asked to predict. Non-ASCII characters were removed here, but included in tweets shown to human annotators. 

\subsection{FTE}
\subsubsection{Topic 0}
\begin{itemize}
	 \item How bad was the blizzard in St. John's,  Newfoundland? Here's what a seniors home looks like the day after.  photo: https://t.co/6n8txqnuWl 
	 \item East End of St. Johns 4 days post blizzard. \#NLwx \#NLtraffic        \#snowmegeddon2020 https://t.co/5s2p9Ivejf 
	 \item Well here's the big mother storm en route. Currently the size of Nova Scotia, nbd. \#nlwx https://t.co/CFi9szzunK 
	 \item INTRUDER: I hope I don't have to go to work on  Monday because I don't remember my password anymore. \#nlwx \#stormageddon2020 https://t.co/BswQppEFVh 
	 \item \#StJohns declares \#StateOfEmergency and \#Newfoundland and \#Labrador get pounded with \#SnowFall with more to come  https://t.co/JNxnIF5mNx 
\end{itemize}
\subsubsection{Topic 1}
\begin{itemize}
	 \item Gentle heart of Jesus \#nlwx https://t.co/cEbvv3it5f 
	 \item This was the moment I fell in love with \#Newfoundland, \#Canada when I first entered the Gros Morne National Park. https://t.co/2mSD79u6qC 
	 \item \#nlwx https://t.co/md4pRSafW5 
	 \item @UGEABC families-send a pic of what youre reading!!! @NLESD @PowersGr6 @AndreaCoffin76 @MrBlackmoreGr1 https://t.co/FwEbRe9YJs 
	 \item INTRUDER: So much for the snow  I was really hoping to make a snowman  Did anyone get their snowman built? \#wheresthesnow https://t.co/FT3IJqIVsS 
\end{itemize}
\subsubsection{Topic 2}
\begin{itemize}
	 \item Garbage collection in the City of St. John's is cancelled for the rest of the week. It will resume on Monday, Jan 2 https://t.co/nvWHGHcrwJ 
	 \item City of St. Johns Snow Clearing on Standby :Due to deteriorating conditions and reduced visibility snow clearing o https://t.co/UC7GV2wFrw 
	 \item Memorial Universitys St. Johns, Marine Institute and Signal Hill campuses will remain closed all day. \#nlwx 
	 \item Current outages. St. Johns area.  https://t.co/wEwJKWKxpF  \#nlwx https://t.co/FsIieBLzGw 
	 \item INTRUDER: \$SIC Sokoman Minerals Provides Winter 2020 Exploration Update at Moosehead Gold Project, Central Newfoundland https://t.co/ma9Ai4PLeE 
\end{itemize}
\subsubsection{Topic 3}
\begin{itemize}
	 \item Not everyone has the funds to go to the grocery store (not just during a SOE) Hats off to the food banks that are o https://t.co/sljOgZZUmf 
	 \item Up to 300 troops from across Canada will be asked to work on the response to the unprecedented \#nlblizzard2020. Gag https://t.co/Yw7jUAWUJE 
	 \item INTRUDER: Beautiful @DowntownStJohns the morning after \#Snowmageddon https://t.co/HS7jhLIt4l 
	 \item So pleased to see the joint efforts of @GovNL with the cities/towns during \#snowmaggedon2020. From calling it a SOE https://t.co/0oKer9CLmo 
	 \item Updated story: Five days is a long time without food for people who can't afford to stock up. Staff at one St. John https://t.co/gNJSXMsv3B 
\end{itemize}
\subsubsection{Topic 4}
\begin{itemize}
	 \item Were quite buried in Paradise right now!    Luckily still have power for now.    Stay safe everyone! \#nlwx https://t.co/sOC7TL4Al7 
	 \item Hoping everyone's pets are safe inside your homes.  \#nlblizzard2020 
	 \item @IDontBlog Yikes!  Hoping everyone there stays safe.... \#nlblizzard2020 \#NLStorm2020 
	 \item @DanKudla The weather channel is calling for snow in Toronto, but nothing like that. Hope youre all safe in Newfou https://t.co/0KN2AD3JrS 
	 \item INTRUDER: BREAKING | Province is calling in the military @NTVNewsNL \#Nlwx https://t.co/ggfOlrbYCz 
\end{itemize}
\subsubsection{Topic 5}
\begin{itemize}
	 \item We've made it through the worst of the storm, but \#yyt's State of Emergency remains in effect. Pls stay inside \&amp; sa https://t.co/YGVxNOEzf1 
	 \item INTRUDER: How COLD is it?  Coby took less than 30 seconds to use the facilities this morning!  \#snowmaggedon2020 \#snowstorm https://t.co/f4zZ38MJol 
	 \item Stay safe St. Johns! \#Newfoundland \#snowmaggedon2020 
	 \item Take your time cleaning this up, Newfoundland friends! It is a brutal amount of snow! Be safe! \#nlwx https://t.co/POwfBYWHFy 
	 \item Meanwhile in ON, 15-25cm w/ 50km wind forecast and asked to stay off streets.  \#nlblizzard2020 https://t.co/wwKVnqmM3F 
\end{itemize}
\subsubsection{Topic 6}
\begin{itemize}
	 \item @VOCMNEWS This is amazing. In the next couple of days we could really get so many of our neighbourhood hydrants dug https://t.co/7Z2TOfwpwB 
	 \item @weathernetwork batteries charging and camera gear drying out after 16 hours shooting in blizzard. @MurphTWN \#nlwx https://t.co/yGUMV93Yrw 
	 \item So we visited friends last night and left at the height of the storm ... somewhere under there are a couple of cars https://t.co/NYHHxXT4EI 
	 \item @KrissyHolmes \#nltraffic  just like any other weekday morning coming out of Cbs. Two solid lines of traffic 
	 \item INTRUDER: @BrianWalshWX so its 7:00 pm , how much more snow potentially will fall before noon tomorrow? \#nlblizzard2020 \#nlwx 
\end{itemize}
\subsubsection{Topic 7}
\begin{itemize}
	 \item @StormchaserUKEU A glimpse into the future @yyt \#nlwx 
	 \item Its official - weve named this storm \#BettyWhiteOut2020 in honour of \#BettyWhitesBirthday \#nlwx 
	 \item Even for just a moment with the front door open, snow is hitting you in the face and the wind is taking your breath https://t.co/pJkpwwqDfF 
	 \item INTRUDER: \#nlwx https://t.co/2mnaF7TkSl 
	 \item Open those curtains, let all the sunshine heat in that you can! Solar gain will help us through. \#nlwx https://t.co/AFAubBvWwg 
\end{itemize}
\subsubsection{Topic 8}
\begin{itemize}
	 \item INTRUDER: Plow came by and then the neighbours started a snow clearing party. So thankful  \#nlblizzard2020 \#nlwx https://t.co/QpIapG1N4Y 
	 \item Y seguimos con la supernevada (fuera de lo comn, tambin hay que decirlo) de \#Newfoundland , en \#Canad ! Laia ll https://t.co/O1adKSXbFa 
	 \item If you have to wear a full snowsuit to go shopping, STAY HOME. You do NOT need to be out. You do NOT need a fondue https://t.co/qig24xkrK5 
	 \item @NewfieScumbag Pretty much... love of God! In case you didnt get the memo, keep your packin vehicle off the roads https://t.co/AJe6RpgWBS 
	 \item So our front door looks like a neat little burrow now, at least. \#nlwx \#Snowmageddon2020 https://t.co/qUCdntRJgE 
\end{itemize}
\subsection{BTM}
\subsubsection{Topic 0}
\begin{itemize}
	 \item THIS!! \#nlwx \#nltraffic https://t.co/fGPl0DDJWF 
	 \item 2/2 During a State of Emergency the public is advised to contact their nearest emergency hospital department for em https://t.co/MTbO1MHF2K 
	 \item there's a house in there somewhere \#snowmaggedon2020 https://t.co/r36lWxJPzd 
	 \item INTRUDER: @PaulDoroshenko Yup, sorry as a proud Newfoundlander living on Vancouver Island for the last ten years they totally https://t.co/eNdNvtnJUI 
	 \item Move over, CBC announcers!  \#NLStorm2020 https://t.co/TTvE0LZgIG 
\end{itemize}
\subsubsection{Topic 1}
\begin{itemize}
	 \item INTRUDER: Snowmageddon 2. The return of the snowstorm! Now playing...\#nlwx \#nlsnowstorm2020 \#snowdoor \#blizzard https://t.co/EvBlOdrg23 
	 \item If you're NOT feelin this? Then you have NO ``PULSE"  @therealbigthump   \#SaturdayMorning  \#Eminem https://t.co/r1orKgKGar 
	 \item Click On Photo To Learn More  \#michelleobama Eminem \#SAvENG Ken Starr \#SexEducation \#nlwx   How Cheap https://t.co/jCYbHGDKM1 
	 \item If you're NOT feelin this? Then you have NO ``PULSE"  @therealbigthump \#SaturdayMorning  \#Eminem \#WomensMarch https://t.co/su8qA0qjP0 
	 \item If you're NOT feelin this? Then you have NO ``PULSE"  @therealbigthump  \#SaturdayMorning  \#Eminem \#WomensMarch https://t.co/pORhRKunhs 
\end{itemize}
\subsubsection{Topic 2}
\begin{itemize}
	 \item Stay safe, Newfoundland. Stay warm.  Stay home.  Stay off the roads.    \#Newfoundland 
	 \item Newfoundland peeps - the CBC in Toronto is warning that Toronto is about to be walloped with snow, and that Envir https://t.co/NmWRu8t6rL 
	 \item INTRUDER: \#NLStorm2020 \#nltraffic  Thanks for coming and doing some cleanup on Gower St right now. Many fans in windows watch https://t.co/FZBsIaN7v1 
	 \item My thoughts go to those who need to serve the communities during \#Snowmageddon in NFLD today.  Stay safe while you do what you need to do. 
	 \item @DrAJHalifax They're getting absolutely walloped!    \#nlstorm 
\end{itemize}
\subsubsection{Topic 3}
\begin{itemize}
	 \item Have woken up ... and grateful that the power is still on in the house.   I'm also grateful that the noise of the w https://t.co/xQjXoz3qED 
	 \item A huge THANK YOU to all of the deputies, officers, troopers, medics, firefighters, dispatchers, plow operators and https://t.co/mCnFbDUHd8 
	 \item Watch what happened at 1:56:51 in @DaHonestyPolicy's broadcast: Come Talk With the Ladies \#DaAngels \#DaNation https://t.co/L4zsDuzTHe 
	 \item What do you expect when graduate with good cgpa don't have a job after graduating, Efcc free guys to hustle jor. Dr https://t.co/DYGlpBBmHe 
	 \item INTRUDER: Weve made the Washington Post \#nlwx https://t.co/sjqdJkR7XL 
\end{itemize}
\subsubsection{Topic 4}
\begin{itemize}
	 \item INTRUDER: Its a brand new day \#nlwx https://t.co/l30WBqLHIE 
	 \item Whos up for a twitter game: give us your best movie title related to \#snowmaggedon2020 pls use \#snowmoviesnl so we can retweet you \#nlwx 
	 \item somebody should make  a video of \#snowmaggedon2020  with Greata saying how dare you 
	 \item Doing my civic duty as a Minnesotan today....storming @Target and battling the masses for the last cart to stock up https://t.co/D72nJ0I3em 
	 \item My  13-yr old got tickets to tomorrow night's @Raptors game for Xmas but will miss b/c all flights are grounded at https://t.co/NsvWNwcIix 
\end{itemize}
\subsubsection{Topic 5}
\begin{itemize}
	 \item Mom look! One of my videos made the @BBCWorld news! Bys, tis a rough way to get famous! \#nlwx \#blizzard2020 https://t.co/Y0jx2pycO7 
	 \item Very cool! \#Newfoundland \$gold \$SIC.B \#mooseheadmadness  https://t.co/YKfIBIMCaA 
	 \item The morning after.   Pictures from my Daughter and my Sister.  Both live in St. John's NL  \#NLWX \#Blizzard \#NL https://t.co/UhtFlddTXn 
	 \item It's \#SaturdayMorning and the storm has died down here in Newfoundland. Still blowing hard out there but not as bad https://t.co/Qkoj6XjBc3 
	 \item INTRUDER: We are still in a SOE - STAY OFF THE ROADS so all of the emergency \&amp; essential workers can do their jobs! Even thou https://t.co/2TSmfKYlnZ 
\end{itemize}
\subsubsection{Topic 6}
\begin{itemize}
	 \item @590VOCM  several reports of very slippery conditions at the intersection of Carrick Drive and Stavanger Drive. Ple https://t.co/pDlCBjvLfi 
	 \item Dave says there's a car off the road and now on top of a snowbank on the ORR WB just before Paradise turnoff \#nltraffic 
	 \item Is  @MaryBrowns  open in Mount Pearl? \#nlwx \#companylunch 
	 \item Power outage here in Bonavista. \#nlwx \#nlblizzard2020 \#snowmaggedon2020 
	 \item INTRUDER: The \#nlwx hashtag is making me really appreciate our mildly inconvenient snowstorm here in Wisconsin. 27 inches of https://t.co/dzV0elcy6y 
\end{itemize}
\subsubsection{Topic 7}
\begin{itemize}
	 \item INTRUDER: @CBCNews \#nlwx \#snowmaggedon2020 might be a while before things return to anything close to normal. Main road throu https://t.co/1vHX2B9oPo 
	 \item MUN's St. John's, Signal Hill campuses and Marine Institute will be closed until Jan. 27  \#nlschools \#YYTSOE \#nlstorm2020 \#nlwx 
	 \item \#SOEYYT remains in place for tomorrow, Jan 23. \#nlwx 
	 \item Bell island ferry update \#nltraffic https://t.co/pTFo4FuxfM 
	 \item MUN closed                                        METROBUS OFF THE ROADS       SCHOOLS CLOSED  https://t.co/G3rvNoXVt5 
\end{itemize}
\subsubsection{Topic 8}
\begin{itemize}
	 \item With a 9:30pm snow total of 20cm, today is \#Gander's snowiest day so far this winter. \#NLWx https://t.co/ZYhyV2Xpsm 
	 \item \#nlblizzard2020 \#nlstorm  look at the winds... the gusts are Cat 4 hurricane equivalent  https://t.co/JjFbiMJAdg 
	 \item 10min avg wind speeds of 132.0km/h with max gust of 167.4km/h through 12:10pm at Green Island, Fortune Bay. \#nlwx https://t.co/GtisE875av 
	 \item INTRUDER: This is Lovely. I have always had a soft spot in my heart for \#Newfoundland and the wonderful people there. https://t.co/CqPyzW3vLH 
	 \item There has been the equivalent of 32.7 mm of precipitation since Fri 04:30 at ``ST JOHNS WEST CLIMATE" \#NLStorm 
\end{itemize}

\end{document}